\documentclass[conference]{IEEEtran}
\IEEEoverridecommandlockouts
\pdfoutput=1
\usepackage{cite}
\usepackage{amsmath,amssymb,amsfonts}
\usepackage{algorithmic}
\usepackage{graphicx}
\usepackage{textcomp}
\usepackage{xcolor}
\usepackage{subcaption}
\usepackage{algorithm}
\usepackage{multirow}
\usepackage{slashbox}

\def\BibTeX{{\rm B\kern-.05em{\sc i\kern-.025em b}\kern-.08em
    T\kern-.1667em\lower.7ex\hbox{E}\kern-.125emX}}
\begin{document}

\title{Ensemble  Neural Networks for Remaining Useful Life (RUL) Prediction
}

\author{
\IEEEauthorblockN{Abhishek Srinivasan}
\IEEEauthorblockA{\textit{Connected Systems}, 
\textit{Scania CV AB}, S\"odert\"alje, Sweden\\
\textit{KTH Royal Institute of Technology}, Stockholm, Sweden\\
\textit{RISE Research Institutes of Sweden}, Stockholm, Sweden\\
abhishek.srinivasan@scania.com}\\
\and
\IEEEauthorblockN{Juan~Carlos Andresen}
\IEEEauthorblockA{\textit{Connected Systems}, 
\textit{Scania CV AB}, S\"odert\"alje, Sweden\\
juan-carlos.andresen@scania.com}\\
\and
\IEEEauthorblockN{Anders Host}
\IEEEauthorblockA{
\textit{RISE Research Institutes of Sweden}, Stockholm, Sweden\\
\textit{KTH Royal Institute of Technology}, Stockholm, Sweden\\
anders.holst@ri.se}\\
}
\maketitle
\begin{abstract}
	A core part of maintenance planning is a monitoring system that provides a good
	prognosis on health and degradation, often expressed as remaining
	useful life (RUL). Most of the current data-driven approaches for RUL
	prediction focus on single-point prediction. These point prediction
	approaches do not include the probabilistic nature of the failure. The
	few probabilistic approaches to date either include the aleatoric
	uncertainty (which originates from the system), or the epistemic
	uncertainty (which originates from the model parameters), or both
	simultaneously as a total uncertainty.  Here, we propose ensemble
	neural networks for probabilistic RUL predictions which considers both
	uncertainties and decouples these two uncertainties. These decoupled
	uncertainties are vital in knowing and interpreting the confidence of
	the predictions. This method is tested on NASA's turbofan jet engine
	CMAPSS data-set. Our results show how these uncertainties can be
	modeled and how to disentangle the contribution of aleatoric and
	epistemic uncertainty. Additionally, our approach is evaluated on
	different metrics and compared against the current state-of-the-art
	methods. 
\end{abstract}

\section{Introduction}
The cost of downtime due to failure and its corresponding unplanned maintenance
is high. A well-planned maintenance strategy can better minimize these failure
occurrences. Predictive maintenance (an advanced maintenance planning strategy)
uses models to monitor the health index of a system to schedule a maintenance.
A popular health index is the Remaining Useful Life (RUL), which is the
effective life left of a component measured in number of operational time, such
as number of cycles, number of hours, or amount of air pumped. The two main
streams of RUL modeling approaches are physics based and data-driven based.
\textbf{Physics based} models are mathematical representations of a system
degradation to predict RUL.  For complex systems, one common method for RUL
modeling is to divide the system into subsystems and recurrently modeling its
sub-components individually \cite{lei2016model}. This process of decomposing
the system into smaller sub-systems and modeling them can be repeated until the
desired level of granularity is reached. This granularity selection also
affects the accuracy of the model (in general, the deeper the level of
granularity, the more accurate the model is). This modeling approach can be
time-consuming and deep domain knowledge about the system and sub-systems is
needed. \textbf{Data-driven} models are modeled using data obtained from the
system.  With the developments in machine learning (ML) the process of
data-driven modeling has become more accurate than ever
\cite{tan2021efficientnetv2}. Motivated by the success of deep learning (DL) in
computer vision and text processing \cite{tan2021efficientnetv2,
zhao2023survey} DL has become mainstream among many researchers within PHM.
Currently, state-of-the-art models take two different directions for RUL
modeling for complex systems; on one hand the inputs are directly mapped onto
the RUL \cite{lstm_method,fan2022trend} and on the other hand, when a health
index is possible to be defined or measured, the modeling is done in a two step
procedure \textit{i}) inputs are mapped onto the health index, \textit{ii}) the
health index is mapped onto the RUL \cite{lstm_prob}.  Despite the good
accuracy of the current approaches using DL \cite{fan2022trend, lstm_method,
lstm_prob}, most of them model point estimates of the RUL without considering
the probabilistic nature of the system and uncertainties in the modeling
\cite{fan2022trend}. 

In general, there are two main sources of uncertainties in the modeling
process; \textit{aleatoric uncertainty} which is originated from the system
failing at different operational times, and \textit{epistemic uncertainty}
which comes from uncertainties of the model parameters, e.g. these model
parameters might change with the quantity of available data. Knowing the source
of the uncertainties gives the possibility of taking better decisions based on
the model predictions \cite{hullermeier2021aleatoric}. For instance, when the
epistemic uncertainties are large the model predictions should not be trusted.
This high epistemic uncertainty strongly indicates that the provided input is
different than the trained data distribution. If the aleatoric uncertainties
dominate, then the uncertainties are inherent to the underlying system (or
quality of data) and cannot be reduced by adding any other source of
information. For industrial applications, being able to distinguish between
these uncertainties can be of much help, i.e., \textit{i)} the aleatoric
uncertainty provides information about the variance in the failure process.
This information can be used to know the amount of risk taken when planning the
maintenance. \textit{ii)} the high epistemic uncertainty indicates regions
where more data collection is needed to enrich model's knowledge. This
distinction gives crucial information to interpret the model output more
accurate in relation to the uncertainties, thus improving the trustworthiness. 

 In this work, we predict the probabilistic estimates; incorporating the
 aleatoric and the epistemic uncertainties by utilizing an ensemble neural
 network. This ensemble based approach is simple, easily parallelizable, and
 well calibrated to reflect real underlying behavior.  Our methods are tested
 on NASA's turbofan jet engine CMAPSS data-set benchmark~\cite{data_ref}. The
 results show the capability of our model approach to provide probabilistic
 estimates and can measure the isolated effect of the aleatoric and epistemic
 uncertainties. 

The paper begins with related work followed by ensemble neural networks for
probabilistic modeling, then we describe the experiments and results. Finally,
we show some advantages of this method and conclude this work.

\section{Related Work}
A number of different authors use neural networks to predict the RUL of a
system. The most common neural network architectures for this application are
Convolution Neural Networks (CNN) and Long Short-term Memory (LSTM).
\cite{lstm_method} use an LSTM network combined with fully connected layers
that take in normalized data and predict RUL.  \cite{fan2022trend} use CNN with
attention mechanism to the predict the RUL along with some interpretability
methods. The aforementioned approaches model for point prediction, our work
aims to model probabilistic predictions incorporating uncertainties.

Some work that considers probabilistic prediction are \cite{lstm_prob};
\cite{muneer2021data}; \cite{nguyen2022probabilistic};
\cite{mitici2023dynamic}.  The work of \cite{lstm_prob} uses three-step model
for probabilistic RUL prediction. The first step is to predict the probability
distribution health index. In the second step, the predicted distribution of
the health index is mapped onto the RUL estimated distribution. The third step
is a correction carried out using LSTMs, this step acts as a re-calibrator for
the prediction. Although the uncertainty estimation on the NN is similar to our
work, one crucial difference between this work and Nemani's work is that our
method is a single-stage prediction where inputs are mapped directly onto the
RUL. This is important in complex systems such as CMAPSS where defining a
health index that is interpretable and observable is difficult or even
impossible.

\cite{mitici2023dynamic} use Monte Carlo dropout approach for probabilistic
predictions and it requires high computation and modeling time compared to our
approach \cite{lakshminarayanan2017simple}.  \cite{nguyen2022probabilistic} use
an approach of modeling which only takes into account uncertainties from the
system and does not model the uncertainties of model parameters. Another
approach by \cite{muneer2021data} where they measure the uncertainties from the
model (epistemic) and don't consider the uncertainties from the system
(aleatoric).

Most of the existing work focuses on modeling point prediction for the RUL and
only a few focus on probabilistic methods. To our knowledge the existing
probabilistic methods either estimate the aleatoric, or the epistemic
uncertainties, or both simultaneously without separating the source of
uncertainties. Our approach models a probabilistic approach that distinguishes
the source of uncertainties.

\section{Methods}
\subsection{Ensemble Neural Networks for Prediction}
\cite{lakshminarayanan2017simple} proposed a novel approach to model both
aleatoric and epistemic uncertainties using deep ensembles probabilistic
networks. Individuals of an ensemble are made of probabilistic neural networks
(PNN). This PNN is a probabilistic model which captures aleatoric uncertainties
from a given data. PNNs work like a neural network with the difference that
they predict the parameters $\theta$ of the assumed distribution
$\Pi(\theta)$
Additionally, epistemic uncertainties are captured by the ensembles, by the
fact that individuals in the ensemble converges to different optimums while
capturing the distribution of the model parameters. During the training
process, the optimizer aims to find parameters for the PNN to maximize the
selected scoring rule.
		
The \textit{Scoring rule} is a function that measures the quality of the
predicted distribution $p_\theta$. The higher the value is, the better the
quality of prediction is. This scoring rule helps to check if the model is
calibrated \textit{i.e.}, the predicted distribution $p_\theta$ reflects the
real distribution $q$, where $\theta$ is the parameter of the assumed
distribution. A well-defined scoring rule should satisfy the following
conditions: \textit{i}) $S(p_\theta,Y|x)<S(q,Y|x)$ and \textit{ii})
$S(p_\theta,Y|x)=S(q,Y|x)$ if and if only $p_\theta(Y|x) = q(Y|x)$. Negative
log likelihood (NLL) and  Brie score are some examples of scoring rules that
satisfy the above properties. 
			
\subsection{Proposed Model Structure}
The proposed model uses a Gaussian distribution $\mathcal{N}(\mu,\sigma)$ as
the assumed distribution $\Pi(\theta)$, where $\mu$ is the mean and $\sigma$ is
the standard deviation. In other words, the distribution of the RUL estimates
is assumed to be Normal distributed. The model architecture consists of $K$
stacks of LSTM layers followed by $L$ fully connected layers which output two
parameter estimates $\hat{\mu}$ and $\hat{\sigma}$. This network is trained
using the NLL of the Gaussian distribution, and the training data is used as
observations on the predicted distribution. The NLL of the $i^{th}$ sample is
given by Eq.~(\ref{equ:NLL_gaussian}). Our modeling approach predicts the RUL
at every time step of the provided window. 	
\begin{equation}
	\label{equ:NLL_gaussian}
	-\log \boldsymbol{p}_{\mu,\sigma}\left(y_i \mid
	\mathbf{x}_i\right)=\frac{\log
	\sigma^2(\mathbf{x}_i)}{2}+\frac{\left(y_i-\mu(\mathbf{x}_i)\right)^2}{2
	\sigma^2(\mathbf{x}_i)}+\text { const }.
\end{equation}
The prediction from $M$ individuals of ensembles is put together by finding the
mean distribution $\mathcal{N}(\hat{\mu}_*,\hat{\sigma}_*)$ 
\begin{align}
	\hat{\mu}_{\ast}&=\frac{1}{M}\sum_{i=1}^{M}\hat{\mu}_i\;,\\
	\hat{\sigma}^2_{\ast}&=\frac{1}{M}\sum_{i=1}^{M}(\hat{\sigma}^2_i +
	\hat{\mu}^2_i) - \hat{\mu}^2_{\ast}.
\end{align}
\subsection{Uncertainty Measures}
As mentioned before the total uncertainty can be split into aleatoric
and epistemic, which can be expressed as $U_{tot}= U_{al} +U_{ep}$.
Aleatoric uncertainty can be measured by the average entropy $H$ of
each prediction, this is $U_{al}=\frac{1}{M}\sum^M_{i=1}
H(\boldsymbol{p}^{(i)})$, where $M$ is the total number of models in
the ensemble, $i$ is an individual in the ensemble and
$\boldsymbol{p}^{(1)}, \dots,\boldsymbol{p}^{(M)}$ are the $M$
predictive distributions of the ensemble.  The total uncertainty
$U_{tot}$ can be calculated as the entropy of the mean prediction,
i.e., $U_{tot}=H(\frac{1}{M}\sum^M_{i=1}\boldsymbol{p}^{(i)})$.
Therefore, $U_{ep}=H(\frac{1}{M}\sum^M_{i=1}\boldsymbol{p}^{(i)}) -
\frac{1}{M}\sum^M_{i=1}
H(\boldsymbol{p}^{(i)})$~\cite{malinin2019ensemble}. By assuming a
Normal distributed variable, i.e., $x \sim \mathcal{N}(\mu,\sigma)$,
the entropy can be expressed as $H=\frac{1}{2} \log
(2\pi\sigma^2)+\frac{1}{2}$. Therefore, $U_{tot}=\frac{1}{2}
\log(2\pi\hat{\sigma}^2_{\ast})+\frac{1}{2}$ and we can write the
\textit{aleatoric} and \textit{epistemic} uncertainties as 
\begin{equation}
	U_{al} \sim \frac{1}{M}\sum_{i=1}^M \log\Big(\hat{\sigma}_i^2\Big)\,,\\	\label{eq.Ual} 
\end{equation}
\begin{equation}
	U_{ep} \sim \log(\hat{\sigma}_{\ast}^2) - \frac{1}{M}\sum_{i=1}^M \log\Big(\hat{\sigma}_i^2\Big)\,.\label{eq.Uep}
\end{equation} 
\section{Experimental Setting}
\subsection{Data}
Our proposed method was tested on NASA's turbofan jet engine CMAPSS
data-set \cite{data_ref}, specifically using FD001 for training and
test sets form FD001, FD002 and FD003 data-sets. These data-sets were
curated for RUL prediction tasks, containing 21 selected signals
collected during different operational cycles until failure. we omitted
sensor signals 1, 5, 10, 16, 18, and 19 as their values are constant in
data-set FD001. We utilize piecewise linear RUL targets; in the initial
stages we assume the RUL to be a constant of value $128$ and linearly
decreasing in the last $128$ cycles, similar to previous approaches
\cite{zhang2019remaining, nguyen2022probabilistic}.
\begin{table}[h] \small  
	\begin{center}  
		\caption{Table summarizing the NASA turbofan jet engine
		data-set. This consists of four data-set with different number
		of units, operating conditions, and fault modes.}
		\label{data_summary_table}
		\begin{tabular}{ l | l | l | l | l   }
			\hline \hline
			\textbf{}	& \textbf{FD001}	&
			\textbf{FD002}   	& \textbf{FD003}    	&
			\textbf{FD004}    \\ 
			\hline \hline
			Train Units & 100 & 260 & 100 & 249 \\ \hline
			Test Units & 100 & 259 & 100 & 249\\ \hline
			Operating Condition & 1 & 6 & 1 & 6\\ \hline
			Fault Modes & 1 & 1 & 2 & 2 \\ \hline
		\end{tabular}
	\end{center}
\end{table}

The data is pre-processed, where the signals are normalized using the $Z$-norm
$x_i^{norm} = (x_i-\mu_x )/\sigma_x$. The normalizing parameters of the train
data are utilized for normalizing the test. Additionally, the sliding window
method is used to generate samples that are used as inputs to the neural
networks. This is typically done by using a window of length $l$ and this
window is moved along time on stride $s$. For this work, the stride $s$ was set
to 1 and the window length $l$ was set to 100.

\subsection{Model}
For reproducibility purposes, the experiments utilized a fixed random seed 237.
Our model uses 2 layers of LSTM layers each with 32 and 16 neurons,
respectively. LSTM layers are followed by 1 dense layer. Our ensemble consists
of 15 models. Train and test split is according to the original data-set. Our
models utilize a batch-size of $32$ and an Adam optimizer with a learning rate
of $\lambda=0.001$, parameters $\beta_1 = 0.9$, and  $\beta_2 = 0.999$. An
early stopping mechanism  monitors loss from epoch $35$ and waits for $3$
epochs to cut off the training when loss continues to increase or at 100
epochs. 

\subsection{Evaluation Metric}
In order to compare against the point prediction methods, we evaluate our
method against the same metrics that are used in point prediction methods. For
this purpose, the mean measure is calculated. Commonly used metric for point
predictions are Root Mean Square Error (RMSE) shown in Eq.~(\ref{rmse}), where
$N$ is the number of samples in the data-set and $\hat{y}$ is the model
prediction. The Score function is shown in Eq.~(\ref{score_fun}) where $a_1$ is
set to $10$ and $a_2$ to $13$ as in \cite{data_ref}.
	
\begin{equation}
	\label{rmse}
	RMSE = \sqrt{ \frac{1}{N} \sum_{i=1}^{N} (\hat{y}_i-y_i)  }\;
\end{equation}

\begin{equation}
	\label{score_fun}
	s=\left\{\begin{array}{l}
		\sum_{i=1}^N e^{-\left(\frac{\hat{y}_i-y_i}{a_1}\right)}-1
		\text { for } (\hat{y}_i-y_i)<0 \\
		\sum_{i=1}^N e^{\left(\frac{\hat{y}_i-y_i}{a_2}\right)}-1 \text
		{ for } (\hat{y}_i-y_i) \geq 0
	\end{array}\right.
\end{equation}

For evaluating the probabilistic predictions, we use the prediction interval
coverage percentage (PICP) and normalized mean prediction interval width
(NMPIW). PICP measures the percent of the prediction which falls within the
bounds given the confidence interval. NMPIW measures the average width of the
bounds, \textit{i.e.}, upper and lower-bound in a possible range of values.
Formulae for PICP and NMPIW are provided in Eq.~(\ref{eq:PICP}) and
Eq.~(\ref{eq:NMPIW}), respectively,
\begin{equation}
	\label{eq:PICP}
	PICP = \frac{1}{N} \sum_{i=1}^N \left\{\begin{array}{l}
		1 \text{ if } y_i \in [U_{\alpha}(\hat{\boldsymbol{p}}_i),
		L_{\alpha}(\hat{\boldsymbol{p}}_i)] \\
		0 \text{ if } y_i \notin [U_{\alpha}(\hat{\boldsymbol{p}}_i),
		L_{\alpha}(\hat{\boldsymbol{p}}_i)]
                \end{array}\right.\;,
        \end{equation}

        \begin{equation}
            \label{eq:NMPIW}
	    NMPIW =  \frac{1}{N(\max{\left\{y\right\}}-\min{\left\{y\right\}})}
		\sum_{i=1}^N
		(U_\alpha(\hat{\boldsymbol{p}}_i)-L_\alpha(\hat{\boldsymbol{p}}_i)),
        \end{equation}
where $\hat{\boldsymbol{p}}_i$ is the estimated distribution by the $i^{th}$
individual in the ensemble. The upper bound $U_{\alpha}(\boldsymbol{p})$ and
lower bound $L_{\alpha}(\boldsymbol{p})$ are calculated based on the confidence
interval $\alpha$ of the distribution $\boldsymbol{p}$. We use a 95\%
confidence interval for our calculations.
\section{Results And Discussion}
\begin{figure}[h]
	\centering
	\includegraphics[scale=0.55]{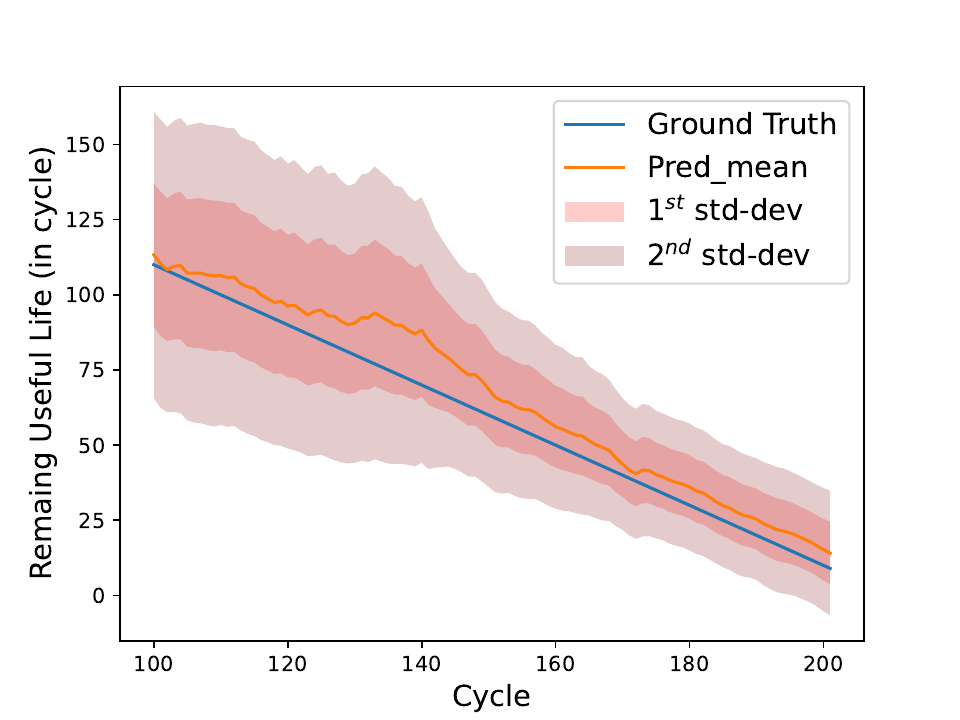}
	\caption{Prediction of unit 34 from the test set in FD001 using model
	trained on train set form FD001. Here the predictions are for the last
	102 window steps.}
	\label{fig:prediction}
\end{figure}
\begin{figure}[h]
	\centering
	\begin{subfigure}[b]{0.3\textwidth}
		\centering
		\includegraphics[width=1.15\textwidth]{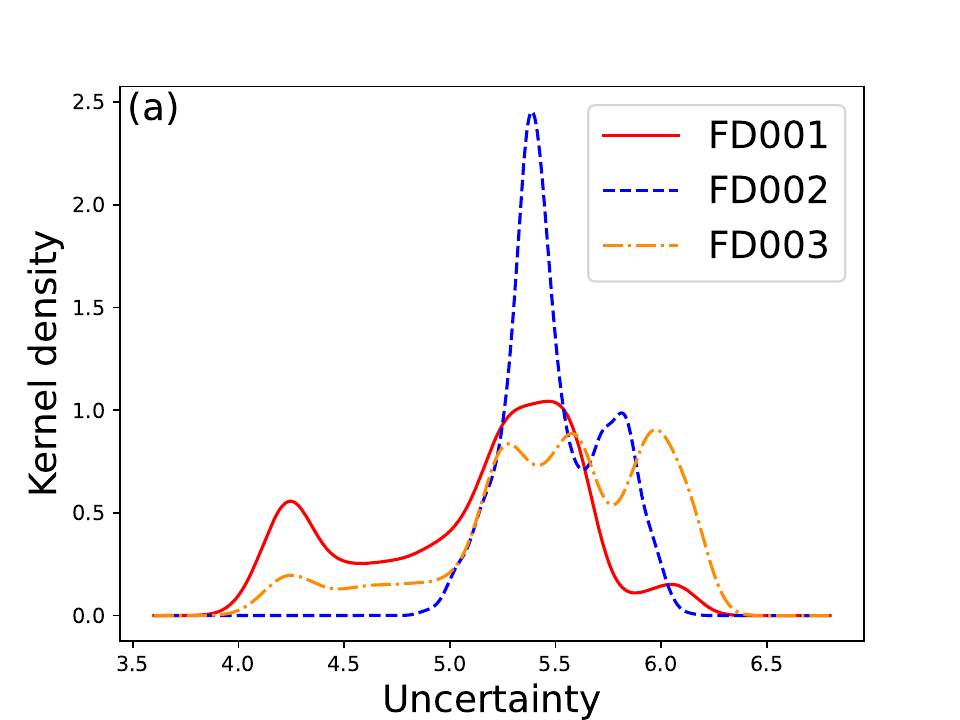}
		\label{fig:al_uncer_hist}
	\end{subfigure}
	\hfill
	\begin{subfigure}[b]{0.3\textwidth}
		\centering
		\includegraphics[width=1.15\textwidth]{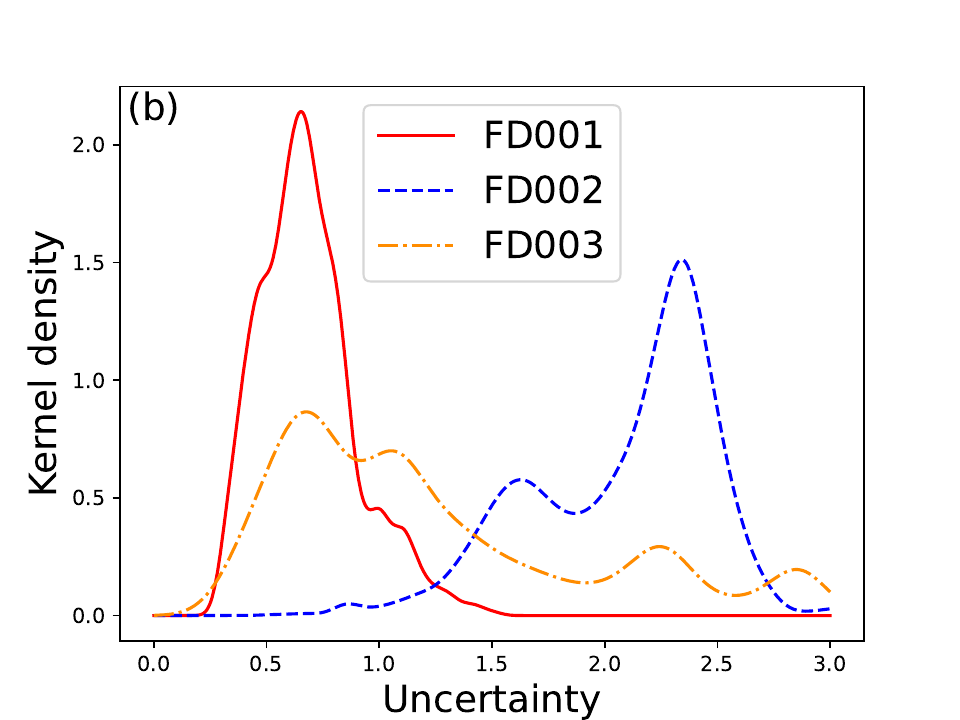}
		\label{fig:epi_uncer_hist}
	\end{subfigure}
	\hfill
	\begin{subfigure}[b]{0.3\textwidth}
		\centering
		\includegraphics[width=1.15\textwidth]{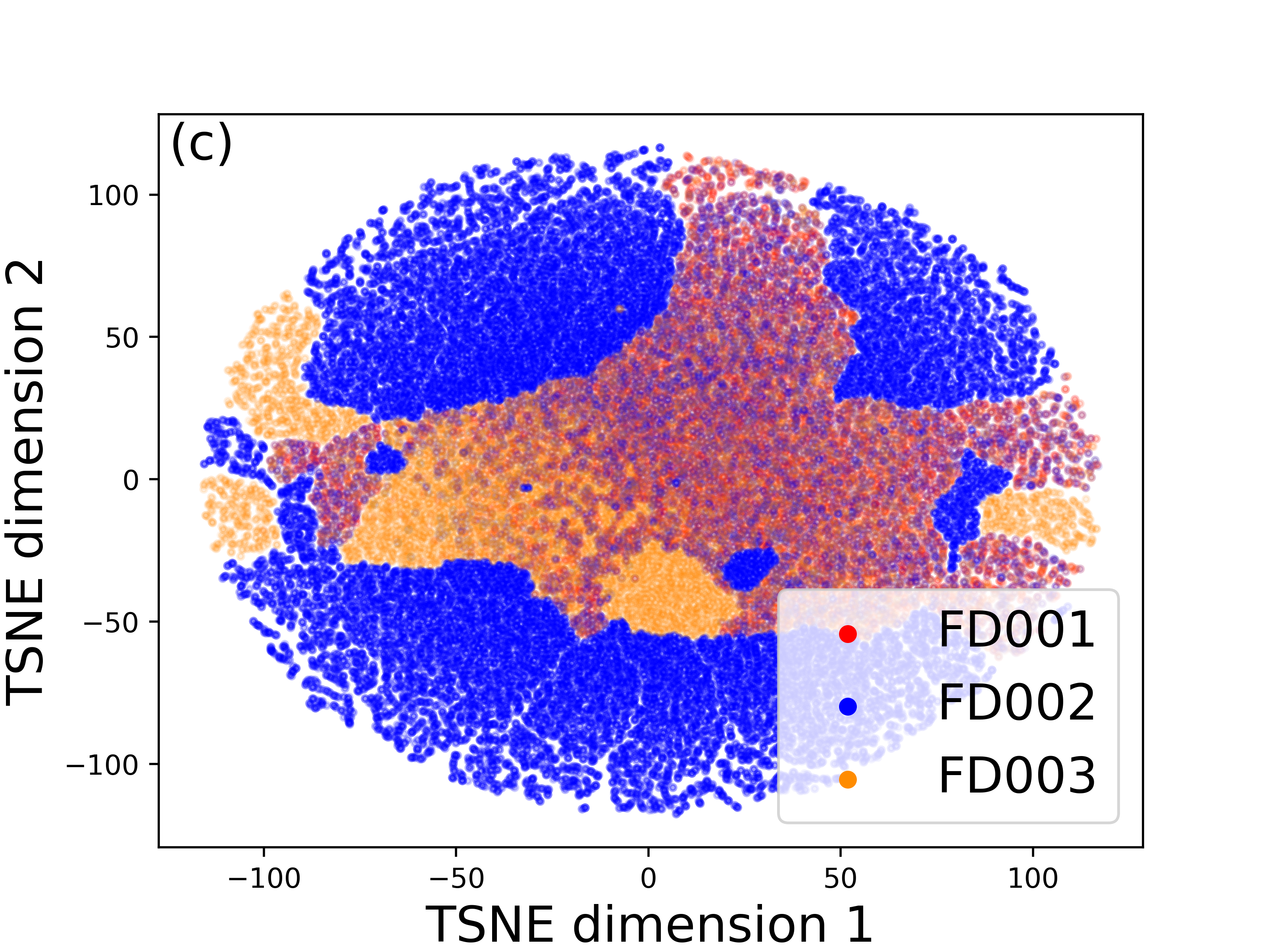}
		\label{fig:tsne}
	\end{subfigure}
	\caption{Kernel density plots of aleatoric uncertainties in (a) and
	epistemic uncertainties in (b) over test sets from folder FD001, FD002,
	and FD003 when predicted over ensemble model trained on FD001. The
	uncertainties of FD001 are plotted in a red solid line, FD002 in the
	dashed blue line, and FD003 in a dash-dotted orange line. (c) shows
	TSNE embedding where projections of data on TSNE dimension 1 and TSNE
	dimension 2. The data from different data-sets is provided in different
	colors red for FD001, blue for FD002, and orange for FD003.}
	\label{fig:three graphs}
\end{figure}
\begin{table*} \small  
	\begin{center}  
		\caption{Table showing the comparison of our  method with
		state-of-the-art methods on point prediction metrics and
		probabilistic metrics. The direction of the arrow indicate what
		makes better model lower or higher. The approaches on the top
		are point prediction methods and the approaches in the bottom
		are probabilistic methods. They are separated by a double
		line.}
		\label{result_table_point_pred_1}
    			\begin{tabular}{ l || l | l || l | l || c }
    				\hline \hline
			&\multicolumn{2}{c||}{
				\begin{tabular}{@{}c@{}}\textbf{Point} \\
					\textbf{Prediction}\end{tabular}} &
					\multicolumn{2}{c||}{\begin{tabular}{@{}c@{}}\textbf{Probabilistic}
					\\ \textbf{Prediction}\end{tabular}} &
					\\
                        \hline
				\textbf{Method}	& \textbf{RMSE}$\downarrow$
				& \textbf{S}$\downarrow$ &
				\textbf{PICP}$\uparrow$ &
				\textbf{NMPIW}$\downarrow$  &
				\textbf{Reference}  \\ \hline \hline
    		        RULCLIPPER & 13.266 & \textbf{216.0} & & &\cite{RULCLIPPER}\\ \hline
    		        MODBNE & 15.039 & 334.2 & & & \cite{MOBDNE} \\ \hline
    				Embed-LR1 & 12.449 & 219.0 & & & \cite{EmbedLR}\\ \hline
    				BiLSTM-ED & 14.741 & 273.0 & & & \cite{BilstED}\\ \hline
    				TSCG & 17.438 & 468.5 & & & \cite{TSCG}\\ \hline
    				SBI-EN & 13.583 & 228.0  & & & \cite{SBIEN}\\ \hline
    				MCLSTM & 13.711 & 315.0  & & & \cite{MCLSTM} \\ \hline
                        Deep LSTM & 16.14 & 338.0 & & & \cite{lstm_method} \\ \hline
    				Trend\_CNN & 13.99 & 336.0 & & & \cite{fan2022trend}\\ \hline \hline
                        MC-dropout & 13.06 & - & - &	-&\cite{mitici2023dynamic}  \\ \hline
                        LSTMBS & 14.481 & 481.1 & 0.960 &	0.377&\cite{LSTMBS}  \\ \hline
    				IESGP & 14.720 & 331.9 & \textbf{0.995} & 0.540 & \cite{IESGP}\\ \hline
    				Lognorm-LSTM(Mean) & \textbf{12.227} & 243.8 & 0.950 & \textbf{0.316} & \cite{nguyen2022probabilistic} \\ \hline
    				\textbf{Our Method (Mean)} & 15.01 & 417.0 & 0.956 & 0.473&\\ \hline 
			\end{tabular}
	\end{center}
\end{table*}
In our modeling process, we train by using a window of 100 time steps and
predict all 100 time steps. Usually, RUL models are evaluated by the
prediction done at the last available time step, therefore we utilize only
the last time step to compare with existing models. Prediction for one
test unit can be seen in the Fig. \ref{fig:prediction}, the mean
prediction follows the ground truth and variance decreases later in the
operational life of this random unit.

We train the ensemble model on the folder FD001 and calculated the aleatoric
and epistemic uncertainty for all the samples in test-sets from folders FD001,
FD002, and FD003. The kernel density estimate of the aleatoric and epistemic
uncertainties are plotted in Fig. \ref{fig:three graphs} (a) and Fig.
\ref{fig:three graphs} (b), respectively.  From Fig. \ref{fig:three graphs}
(b), it is clear that the epistemic uncertainties for the samples from FD002
are high compared to the samples in FD001.  This high uncertainty indicates
that the model has not been trained on the data distribution of FD002 and
should not be trusted (\textit{i.e.}, re-training needed for this data-set). In
the case of FD003 the ensemble model has an epistemic uncertainty that is
closer to the FD001, indicating that the prediction can be trusted but are not
as good as for FD001 and data-distribution is closer to FD001. To further
analyze the epistemic uncertainty and how this reflects on the difference in
data distribution of the different data-sets (\textit{i.e.}, FD001, FD002 and
FD003), we plot in Fig. \ref{fig:three graphs} (c) the T-distributed stochastic
neighbor embedding (TSNE), a dimensionality reduction technique on the
data-space of FD001, FD002 and FD003. This visualization shows the data
embedding of the different data-sets, one can see that the FD001 are subsets of
FD002, and  that FD003 is majorly a sub-set of FD001 with minor exceptions that
can be seen on the left boundaries. Fig.~\ref{fig:three graphs} (c) confirms
our interpretation of the epistemic uncertainty in data-set FD002 and FD003. 

In  Fig. \ref{fig:three graphs} (a) we see that the aleatoric uncertainties lie
in the same region for all 3 data-sets. This indicates that the uncertainties
coming from the system are similar in the three data-sets.  This is because the
model was trained to predict the aleatoric uncertainties ($\sigma$ of the
estimates) of FD001 and therefore model predicates aleatoric uncertainties in
the same region as FD001. These uncertainties can only be trusted when the
epistemic uncertainties are low. These aleatoric uncertainties are due to
inherent characteristics of data and can not be reduced by any means.

Finally, to compare against the existing state-of-the-art point-prediction
approaches, we evaluated our approach using point-prediction and probabilistic
metrics. The comparison is shown in Table. \ref{result_table_point_pred_1}. In
this work, the focus is on how to include probabilistic prediction in RUL
modeling and use a simple LSTM model for RUL predictions. From the table, we
see that our simple RUL-LSTM compares well with state-of-the-art point
prediction models. Moreover, our probabilistic approach can be easily
implemented in the best performing RUL predictive models.
\section{Conclusion}
To summarise, we proposed an ensemble LSTM neural network for probabilistic
prediction to incorporate both aleatoric and epistemic uncertainties for RUL
prediction. This approach is tested on NASA's turbofan jet engine CMAPSS
data-set.  Our results show how epistemic and aleatoric uncertainties can be
added to RUL predictions. The knowledge of the uncertainties, especially the
epistemic uncertainty, allows us to estimate the ensemble model prediction
confidence on a given data-set. If the epistemic uncertainty is large, then it
is a strong indication that the ensemble model has not seen this data before
and needs to be re-trained for this data-set. This ensemble probabilistic
approach is simple to implement on already existing RUL point-predictions,
which would significantly improve trust and transparency to current
state-of-the-art predictions.

Further work could explore methods for the selection of optimal distribution in
place of Gaussian distribution based on the data and could perform further
tests to understand the effect of number of models in the ensemble.
\section*{Acknowledgment}
This work is supported by VINNOVA FFI under the contract 2020-05138. We thank
Kuo-Yun Liang for helping us by reviewing this work. Finally, thanks to Scania
CV AB for supporting this research project.
\bibliographystyle{IEEEtran}
\bibliography{IEEEabrv,references}

\end{document}